\documentclass[10pt,twocolumn,letterpaper]{article}
\usepackage{wacv}
\usepackage{iftex}
\ifPDFTeX
  \usepackage{times}
\else
  \usepackage{fontspec}
  \IfFontExistsTF{Times New Roman}{\setmainfont{Times New Roman}}{\setmainfont{Liberation Serif}}
  \IfFontExistsTF{Arial}{\setsansfont{Arial}}{\setsansfont{Liberation Sans}}
\fi
\usepackage{graphicx}
\usepackage{amsmath,amssymb}
\usepackage{booktabs}
\usepackage{multirow}
\usepackage{makecell}
\usepackage{xspace}
\usepackage{microtype}
\usepackage{placeins}
\usepackage{float}
\makeatletter
\newcommand{\includenativegrouppdf}[2][]{%
  \begingroup
  \let\Gread@transgrouptrue\Gread@transgroupfalse
  \includegraphics[#1]{#2}%
  \endgroup
}
\makeatother
\definecolor{wacvblue}{RGB}{0,114,178}

\usepackage{array}
\usepackage[pass]{geometry}
\newcommand{\suppfigurewidth}{0.92\textwidth}
\newcommand{\suppcomparisonwidth}{0.92\textwidth}

\usepackage[breaklinks,colorlinks,allcolors=wacvblue]{hyperref}

\newcommand{\method}{\mbox{\textnormal{\textsc{DroneSegEdit}}}\xspace}

\newcommand{\editor}{\mbox{LongCat-Image-Edit}\xspace}

\newcommand{\paneltag}[1]{{\sffamily\bfseries (#1)}}
\newcommand{\captionpanel}[1]{(#1)}

\def\wacvPaperID{1705}
\def\confName{WACV}
\def\confYear{2027}
\hypersetup{pdftitle={DroneSegEdit: Adapting Pretrained Generative Editors for UAV Segmentation with Language-Grounded Verification},pdfauthor={Dehua Chen, Long Xin, Jiangtao Wei},pdfsubject={Accepted at WACV 2027; author-accepted manuscript with appendix}}

\title{DroneSegEdit: Adapting Pretrained Generative Editors for\\UAV Segmentation with Language-Grounded Verification}
\author{Dehua Chen \qquad Long Xin \qquad Jiangtao Wei\thanks{Corresponding author: Jiangtao Wei.}\\[7pt]
Fudan University, China\\[2pt]
{\small\texttt{\href{mailto:26113090032@m.fudan.edu.cn}{26113090032@m.fudan.edu.cn}\quad
\href{mailto:longxin25@m.fudan.edu.cn}{longxin25@m.fudan.edu.cn}\quad
\href{mailto:jtwei@fudan.edu.cn}{jtwei@fudan.edu.cn}}}}

\begin{document}
\AddToShipoutPicture*{%
  \AtPageLowerLeft{%
    \put(\LenToUnit{0.8125in},\LenToUnit{0.35in}){%
      \parbox[b]{6.875in}{\fontsize{8}{9.5}\selectfont\color{black}%
      \textcopyright\ IEEE.
      Personal use of this material is permitted. Permission from IEEE must be
      obtained for all other uses, in any current or future media, including
      reprinting/republishing this material for advertising or promotional purposes,
      creating new collective works, for resale or redistribution to servers or
      lists, or reuse of any copyrighted component of this work in other works.}%
    }%
  }%
}

\maketitle

\begin{abstract}
Recent work has shown that pretrained image generators can support diverse visual tasks.  Yet directly applying them to language-guided unmanned aerial vehicle (UAV) segmentation exposes a key limitation: a plausible mask may correspond to the wrong instance among small and repeated objects.  We call this mismatch between mask rendering and expression-level target selection the \emph{Grounding--Rendering Gap}.  We present \method, which combines RGB proposal rendering with language-grounded, proposal-constrained verification to select supported foreground while retaining rendered geometry.  \method adapts the open-weight \editor model using task-specific LoRA and fixed color-to-label decoding.  For referring segmentation, a Language-Grounded Spatial Verifier (LGSV) reads hierarchical states from the frozen multimodal large language model backbone to derive expression-conditioned spatial evidence, which is mapped back to the rendered proposal without introducing foreground outside it.  On RIS-LAD, \method reaches a new state of the art at 52.78 mIoU, 7.16 points above the previous best.  It remains competitive overall on UAVid-RIS and VDD-RIS, with its clearest class-wise gains on compact, application-relevant categories, notably vehicles and humans.  A separate verifier-free experiment extends the RGB interface to full-scene parsing: a task-specific adapter jointly renders all classes from one task-level instruction without enumerating class names or palette colors.
\end{abstract}

\begin{figure*}[!t]
  \centering
  \begin{minipage}{0.94\textwidth}
    \centering
    \includenativegrouppdf[width=\linewidth]{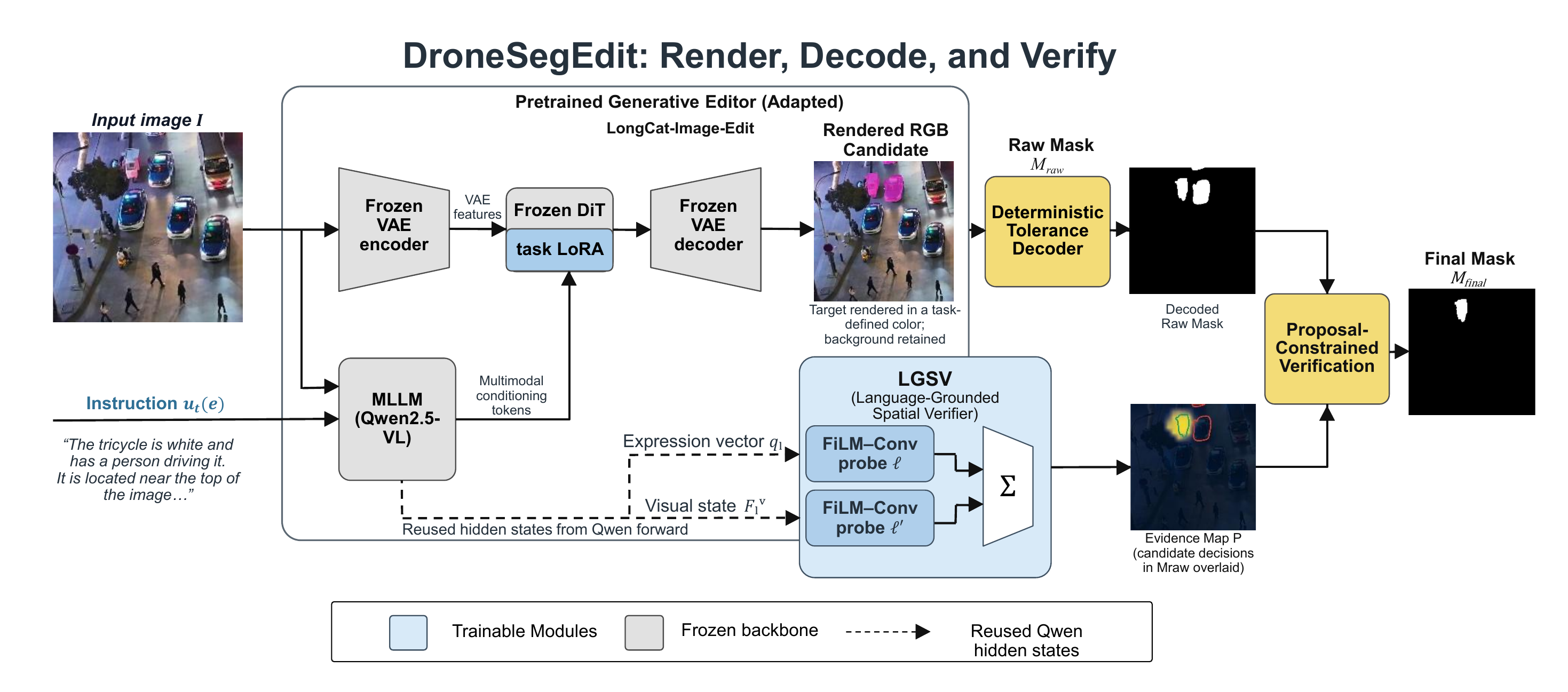}\\[-1mm]
    \paneltag{a}\ {\sffamily\bfseries Overall framework}
  \end{minipage}\\[0.25mm]
  \begin{minipage}{0.94\textwidth}
    \centering
    \includenativegrouppdf[width=\linewidth,trim=0 110bp 0 0,clip]{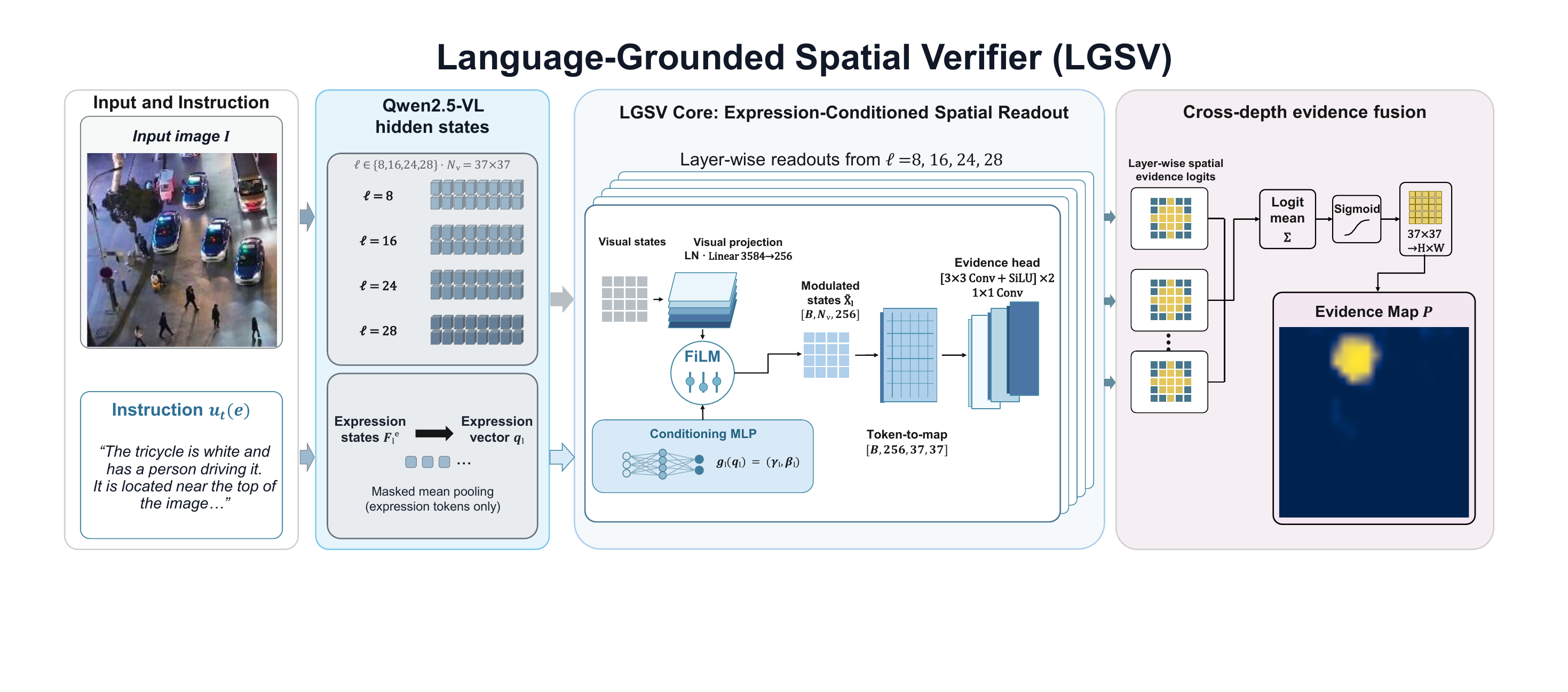}\\[-1mm]
    \paneltag{b}\ {\sffamily\bfseries LGSV architecture}
  \end{minipage}
  \caption{\textbf{DroneSegEdit architecture.}
  \captionpanel{a} The adapted editor renders an RGB proposal, deterministic decoding yields $M_{\mathrm{raw}}$, and LGSV removes expression-unsupported regions. Green and red contour overlays mark retained and rejected candidates, respectively.
  \captionpanel{b} Four FiLM-conditioned probes reuse frozen Qwen2.5-VL states from the editor's conditioning forward pass and fuse complementary spatial-semantic cues into $P$. The editor supplies proposal geometry, while LGSV supplies expression-conditioned support.}
  \label{fig:overview}
\end{figure*}

\section{Introduction}

Referring segmentation converts a natural-language expression into a pixel-level mask.  Unmanned aerial vehicle (UAV) imagery supports civilian applications such as traffic and infrastructure monitoring, where reliable perception of vehicles and people is essential~\cite{lyu2020uavid,cai2025vdd}.  Yet these application-relevant targets often occupy few pixels and occur in dense groups of visually similar instances.  A referring expression may distinguish the intended instance only through subtle spatial or contextual cues.  A successful system must therefore answer two coupled questions: which instance is being referred to, and which pixels belong to that instance.

Most referring systems rely on discriminative cross-modal decoders or multimodal reasoning followed by promptable segmentation~\cite{lai2024lisa,chen2024sam4mllm,lv2026resap,kirillov2023sam,nguyen2025vatex,ito2025feature}.  In parallel, image-space generalist models show that heterogeneous visual tasks can share an RGB prediction interface~\cite{wang2023painter,wang2023seggpt,geng2024instructdiffusion,qi2024unigs,zhao2025diception}, while recent evidence further indicates that generative pretraining provides transferable visual priors~\cite{gabeur2026visionbanana}.  For UAV perception, this formulation can retain fine spatial structure, produce visually inspectable predictions, and provide a common output space across heterogeneous pixel-level tasks.  Its use for language-guided aerial segmentation, however, remains largely unexplored.

We adopt the open-weight, instruction-guided \editor~\cite{longcat2025}.  Its backbone combines Qwen2.5-VL for multimodal instruction understanding with a 6B-parameter diffusion transformer~\cite{peebles2023dit} for image generation.  It reports leading performance among open-weight image-editing models on standard benchmarks, providing a strong generative prior for UAV segmentation.  We keep the pretrained editor frozen and adapt it with lightweight LoRA updates.  A unified RGB output interface encodes segmentation targets as task-defined colors and recovers discrete masks through fixed color-to-label decoding.  For referring segmentation, the referred region is rendered in a specified color while the remainder of the input image is preserved.

Pretrained generative editors offer useful visual priors for rendering spatially coherent RGB masks, but plausible mask generation does not guarantee expression-level target selection among small, repeated, or visually similar UAV instances.  We call this mismatch between rendered regions and expression-supported regions the \emph{Grounding--Rendering Gap}.  \method addresses it by rendering an RGB proposal and then applying language-grounded, proposal-constrained verification: LGSV estimates expression-conditioned support, and the final mask remains restricted to the rendered proposal.  The combination preserves rendered boundaries while using coarse language evidence to select foreground; matched-candidate controls test this complementary role (Table~\ref{tab:ablation}).

The Language-Grounded Spatial Verifier (LGSV; Fig.~\ref{fig:overview}) reads four hierarchical layers of the editor's frozen Qwen2.5-VL backbone to check whether proposed regions match the referring expression.  This separates target selection from RGB generation while keeping the rendered regions as the spatial reference.  Given a palette-decoded candidate $M_{\mathrm{raw}}$ and LGSV support map $P$, the shared verification operator is
\begin{equation}
\begin{aligned}
M_{\mathrm{final}}
&=\mathcal V_t(M_{\mathrm{raw}},P)\\
&:=M_{\mathrm{raw}}\cap\mathcal S_t(M_{\mathrm{raw}},P)
\subseteq M_{\mathrm{raw}},
\end{aligned}
\label{eq:subset}
\end{equation}
Equation~\ref{eq:subset} makes verification one-sided: rendering supplies candidate geometry and LGSV removes unsupported foreground without synthesizing new regions.  Single-instance and region-union tasks share this operator; only the support-set definition follows the required output topology.

Our main contributions are:
\begin{itemize}
  \item We repurpose a pretrained open-weight generative editor for UAV RGB mask rendering.  A unified RGB output interface with fixed color-to-label decoding supports both referring segmentation and full-scene parsing through task-specific LoRA adaptation over the same frozen editor.
  \item We introduce LGSV and proposal-constrained verification to resolve expression-dependent target selection using frozen hierarchical multimodal states while restricting corrections to editor proposals.
  \item Across three UAV referring benchmarks, \method establishes a new state of the art on RIS-LAD while remaining competitive overall on UAVid-RIS and VDD-RIS, with its clearest class-wise gains on compact vehicle and human targets.  Matched-candidate direct-generation and simple proposal-filtering controls isolate the complementary value of language-grounded selection.  A separate verifier-free experiment extends the RGB interface to full-scene UAV parsing from one task-level instruction without per-class prompt enumeration.
\end{itemize}

\section{Related Work}

\paragraph{Referring and aerial segmentation.}
Conventional RIS uses cross-modal decoders to predict the requested mask~\cite{hu2016ris,ding2021vlt,li2021reftr,yang2022lavt,wang2022cris}.  Prompt- and mask-grounding variants refine instance contrast or pixel--language alignment~\cite{shang2024promptris,chng2024maskgrounding}; LISA, SAM4MLLM, and ReSaP route multimodal reasoning to dedicated mask backbones~\cite{lai2024lisa,chen2024sam4mllm,lv2026resap}.  UAV and remote-sensing imagery further introduces arbitrary orientation, scale variation, compact targets, and repeated instances~\cite{yuan2024rrsis,liu2024rmsin,lei2025fianet,chen2025rsrefseg,shi2025mafn,ye2026rislad,li2025aeroreformer}.  In contrast, our adapted editor supplies proposal geometry and language evidence selects within that candidate rather than redrawing its boundaries.

\paragraph{Generative segmentation and image-space interfaces.}
Painter, SegGPT, and InstructDiffusion formulate heterogeneous visual tasks as image-space prediction~\cite{wang2023painter,wang2023seggpt,geng2024instructdiffusion}.  GSS, GenMask, and recursive noise diffusion study direct mask or semantic-map generation, including aerial views~\cite{chen2023gss,yang2026genmask,kolbeinsson2024recursive}.  Concurrent work evaluates released image editors, including LongCat-Image-Edit, as zero-shot dense predictors on generic benchmarks without task-specific adaptation, and excludes instance segmentation because of inconsistent instance-level outputs~\cite{liu2026editors}.  In contrast, we study parameter-efficient adaptation for UAV segmentation: background-preserving RGB rendering supplies a candidate, and language-grounded verification addresses the Grounding--Rendering Gap for compact and repeated referents.  The separate dense experiment uses a fixed task-level instruction rather than a per-image ground-truth class list.

\paragraph{Generative-feature readouts and language grounding.}
CLIP and DenseCLIP transfer image--text alignment to dense prediction~\cite{radford2021clip,rao2022denseclip}.  ODISE and VPD reuse diffusion representations for perception~\cite{xu2023odise,zhao2023vpd}, whereas DMP adapts a pretrained diffusion model for dense prediction~\cite{lee2024dmp}.  Readout Guidance controls sampling through a learned readout~\cite{luo2024readout}.  Related PEFT methods adapt vision, diffusion, and geospatial models with small trainable modules~\cite{chen2022adaptformer,xie2023difffit,scheibenreif2024geoadapt}.  LGSV instead reads hierarchical Qwen states from the editor's existing conditioning forward pass and supplies expression-conditioned evidence within the rendered candidate; it neither reruns nor controls sampling.  Dense parsing uses a closed-set, task-specific adapter.

\section{Method}

\subsection{Unified RGB Rendering}

Let $t\in\mathcal T$ index a UAV task, $I\in[0,255]^{H\times W\times3}$ an input image, $e$ a referring expression when present, $\Omega$ the $H\times W$ pixel lattice, and $Y_t$ the target.  Referring tasks form $\mathcal T_{\rm ref}=\mathcal T_{\rm single}\cup\mathcal T_{\rm union}$; dense semantic tasks form $\mathcal T_{\rm sem}$.  We freeze the pretrained parameters $\theta_0$ and learn standard task-specific low-rank updates $W_j^{(t)}=W_j+\Delta W_j^{(t)}$, with $\operatorname{rank}(\Delta W_j^{(t)})\le r_j$~\cite{hu2022lora}; the factorization and injection details are provided in the appendix.  Let $\mathbf c_{\rm ref}^{(t)}$ denote the task-defined reference color for referring segmentation and $\{\mathbf c_k^{(t)}\}_{k\in\mathcal C_t}$ the palette over the semantic label set $\mathcal C_t$.  The common RGB supervision canvas is
\begin{equation}
Y_{\mathrm{rgb}}^{(t)}(x)=
\begin{cases}
\mathbf c_{\rm ref}^{(t)}, & t\in\mathcal T_{\rm ref},\ Y_t(x)=1,\\
I(x), & t\in\mathcal T_{\rm ref},\ Y_t(x)=0,\\
\mathbf c_{Y_t(x)}^{(t)}, & t\in\mathcal T_{\rm sem}.
\end{cases}
\label{eq:target}
\end{equation}
Thus, the target for referring segmentation is a background-preserving edited photograph, whereas the target for dense semantic segmentation is a complete palette label image; in both cases, the model predicts three-channel RGB rather than mask logits.  The editor learns the color--label correspondence implicitly from RGB supervision, while the fixed palette is used to encode targets and decode predictions.  For referring tasks, the instruction $u_t(e)$ casts prediction as sparse editing: render only the referred region in the task-defined color and preserve all remaining pixels.  Let $\Delta_t=\{\Delta W_j^{(t)}\}_j$ collect the task adapter.  The adapted editor $G$ renders
\begin{equation}
I_g=G_{\theta_0,\Delta_t}(I,u_t(e)),\qquad
M_{\mathrm{raw}}=\mathcal D_{\mathrm{pal}}^{(t)}(I_g).
\label{eq:render}
\end{equation}
The fixed decoder $\mathcal D_{\mathrm{pal}}^{(t)}$ uses one tolerance around the reference-color region for all samples.  $M_{\mathrm{raw}}$ is the rendered candidate before verification.  The background-preserving RGB output $I_g$ can also be inspected directly before decoding.

For dense semantic segmentation, a task-specific adapter receives the fixed instruction ``Convert the full image into an RGB semantic label map.'' and renders all classes jointly in one generation.  The instruction enumerates neither class names nor palette colors; paired RGB targets teach their correspondence.  Nearest-color decoding gives
\begin{equation}
\widehat Y_t(x)=\arg\min_{k\in\mathcal C_t}
\lVert I_g(x)-\mathbf c_k^{(t)}\rVert_2^2.
\label{eq:vdd-palette}
\end{equation}
All tasks therefore share an RGB output interface over the same frozen pretrained editor.  Dense VDD uses the verifier-free palette path to render all classes jointly in one generation from a single task-level instruction.

\subsection{Language-Grounded Spatial Evidence}

The frozen Qwen2.5-VL backbone~\cite{bai2025qwen} processes $(I,u_t(e))$ in the same conditioning forward pass used by the editor.  LGSV taps its hierarchical hidden states and pools the token positions corresponding to the referring expression $e$, as indicated by $m_n$; no additional Qwen pass is required during integrated inference.  At layer $\ell\in\mathcal L=\{8,16,24,28\}$, LGSV reads visual states $F_\ell^v\in\mathbb R^{N_v\times d_\ell}$ and expression-token states $F_\ell^e\in\mathbb R^{N_e\times d_\ell}$.  The image-contextualized expression state and FiLM-modulated visual grid are
\begin{equation}
\begin{aligned}
q_\ell
&=\frac{\sum_n m_nF_{\ell,n}^e}{\sum_n m_n},
\qquad (\gamma_\ell,\beta_\ell)=g_\ell(q_\ell),\\
U_\ell
&=\operatorname{Grid}\!\bigl(\phi_\ell(\operatorname{LN}(F_\ell^v))\bigr),\\
\widetilde U_\ell
&=U_\ell\odot(1+\kappa\gamma_\ell)+\kappa\beta_\ell.
\end{aligned}
\label{eq:film}
\end{equation}
Here $\phi_\ell$ projects visual tokens, $g_\ell$ generates the channel-wise FiLM parameters $(\gamma_\ell,\beta_\ell)$, and $\operatorname{Grid}$ restores the token layout.  The fixed scale $\kappa$ controls the residual modulation magnitude~\cite{perez2018film}.  Two convolution--SiLU blocks and a pointwise head $h_\ell$ produce logits $Z_\ell=h_\ell(\widetilde U_\ell)$.  Uniform logit averaging followed by bilinear upsampling yields
\begin{equation}
P=\operatorname{Up}_{H,W}\!\left(
\sigma\!\left(\frac14\sum_{\ell\in\mathcal L}Z_\ell\right)\right),
\label{eq:evidence}
\end{equation}
where $P\in[0,1]^{H\times W}$ is a relative language-conditioned support map, not a calibrated posterior or an independent boundary mask.  LGSV is a readout over the frozen backbone.  This separates RGB rendering from expression verification: the editor produces an inspectable RGB candidate, and LGSV removes foreground unsupported by the expression.

The two branches are deliberately asymmetric.  The editor predicts in image space and supplies coherent proposal contours, whereas Qwen evidence is produced on a $37\times37$ token grid.  Although language-conditioned, this grid is too coarse to recover thin or irregular boundaries independently.  Treating $P$ as the final mask would therefore require the readout to solve boundary reconstruction.  We instead use $P$ only to select support within $M_{\rm raw}$, enforcing $M_{\rm final}\subseteq M_{\rm raw}$ and preventing verification from introducing foreground not proposed by the editor.

Every active depth is supervised before aggregation.  Let $O_\ell=\mathbf 1[\operatorname{AdaMaxPool}_\ell(Y_t)>0]$ denote target occupancy on the $\ell$-th evidence grid, and let $\delta_t=\mathbf 1[t={\rm RIS\mbox{-}LAD}]$.  Full-grid BCE labels non-target cells as negatives under expression conditioning, while Dice penalizes excessive predicted support.  For single-instance RIS-LAD, center-cell supervision additionally encourages the evidence peak to fall near the target centroid:
\begin{equation}
\begin{gathered}
\mathcal L_{\rm LGSV}^{(t)}
=\frac1L\sum_{\ell\in\mathcal L}\!\Big[
\lambda_b\mathcal L_{\rm BCE}^{(\ell)}
+\lambda_d\mathcal L_{\rm Dice}^{(\ell)}\\[-1pt]
+\delta_t\lambda_c\mathcal L_{\rm center}^{(\ell)}\Big].
\end{gathered}
\raisetag{-4pt}
\label{eq:loss}
\end{equation}
For region-union tasks, where multiple same-class regions may be valid, $\delta_t=0$ and only the center-cell term is disabled; the full-grid BCE and Dice losses remain active.  All fixed coefficients and target constructions are given in the appendix.

\subsection{Proposal-Constrained Verification}

LGSV applies the set operator $\mathcal V_t$ in Eq.~\ref{eq:subset} to both single-referent and region-union outputs.  Let $\Omega=\{1,\ldots,H\}\times\{1,\ldots,W\}$ be the image lattice.  The task topology changes only the construction of the evidence support set $\mathcal S_t\subseteq\Omega$.

For component competition, decompose $M_{\mathrm{raw}}=\bigcup_{i=1}^NC_i$ into 8-connected components and define $\mathcal E=\{i:|C_i|\ge A_{\min}\}$.  Each eligible component receives the mean LGSV evidence within its rendered contour,
\begin{equation}
s_i=\frac{1}{|C_i|}\sum_{x\in C_i}P(x),
\qquad i\in\mathcal E.
\label{eq:score}
\end{equation}
For a single referent, retain eligible components whose evidence is within a factor $\rho$ of the strongest component.  For a region union, dilate the high-evidence seed set.  When $|\mathcal E|>1$, define
\[
\begin{aligned}
\mathcal K_\rho
&=\left\{i\in\mathcal E:s_i\ge\rho\max_{j\in\mathcal E}s_j\right\},\\
\mathcal H_t
&=\{x\in\Omega:P(x)\ge\tau_t\}.
\end{aligned}
\]
Let $\mathcal B_d(A)=\bigcup_{x\in A}\{y\in\Omega:\lVert y-x\rVert_1\le d\}$ denote $\ell_1$ dilation.  The two support constructions are
\begin{equation}
\begin{aligned}
\mathcal S^{\rm single}
&=\bigcup_{i\in\mathcal K_\rho} C_i,
&&|\mathcal E|>1,\\[2pt]
\mathcal S^{\rm union}
&=\mathcal B_{d_t}(\mathcal H_t).
\end{aligned}
\label{eq:support}
\end{equation}
When $|\mathcal E|\leq1$, the single-referent branch sets $\mathcal S_t=\Omega$, leaving the prediction unchanged.  Otherwise, $\mathcal S_t=\mathcal S^{\rm single}$ for single-referent tasks and $\mathcal S_t=\mathcal S^{\rm union}$ for region-union tasks.  Here $\rho$, $\tau_t$, and $d_t$ are fixed on validation data; all constants are listed in the appendix.

The upsampled $37\!\times\!37$ expression-conditioned grid can assign spatially diffuse responses to visually similar instances.  The RIS-LAD rule therefore compares mean evidence across rendered components for relative proposal selection.

The single-referent rule performs competition among eligible rendered components.  Averaging within each contour normalizes for component area, so ranking reflects mean expression support rather than the total evidence accumulated by a large region.  The relative criterion can retain several similarly supported components.  For region unions, thresholding and dilation instead construct spatial neighborhoods around multiple evidence seeds before intersection with the rendered foreground.

\section{Experiments}

\begin{table*}[!t]
\centering
\caption{\textbf{RIS-LAD test results} (\%). The Qwen3-VL SAM4MLLM configuration follows the SFT baseline in ReSaP~\cite{lv2026resap}; \editor and \method share rendered candidates. Best/second-best values are bold/underlined.}
\label{tab:rislad-main}
\footnotesize
\renewcommand{\arraystretch}{1.08}
\setlength{\tabcolsep}{2.4pt}
\begin{tabular}{@{}llrrrrrrr@{}}
\toprule
Method & Core backbone(s) & P@0.5 & P@0.6 & P@0.7 & P@0.8 & P@0.9 & mIoU & oIoU\\
\midrule
\multicolumn{9}{@{}l}{\emph{Traditional RIS methods}}\\
LAVT~\cite{yang2022lavt} & Swin-B/BERT & 31.67 & 27.04 & 23.02 & 17.55 & 8.58 & 30.14 & 41.97\\
FIANet~\cite{lei2025fianet} & Swin-B/BERT & 40.21 & 35.22 & 29.93 & 23.71 & 13.21 & 37.44 & 43.39\\
RMSIN~\cite{liu2024rmsin} & Swin-B/BERT & 43.36 & 38.11 & 32.36 & 25.08 & 13.75 & 39.60 & 48.82\\
RSRefSeg~\cite{chen2025rsrefseg} & SigLIP/SAM & 44.73 & 40.21 & 36.30 & 30.91 & 21.46 & 41.16 & 47.71\\
SAARN~\cite{ye2026rislad} & Swin-B/BERT & 45.02 & 39.34 & 33.37 & 26.24 & 15.31 & 41.67 & 49.60\\
\midrule
\multicolumn{9}{@{}l}{\emph{MLLM-based segmentation models}}\\
LISA~\cite{lai2024lisa} & LLaVA-7B/SAM & 33.20 & 28.17 & 25.57 & 19.62 & 6.28 & 28.50 & 38.39\\
SAM4MLLM~\cite{chen2024sam4mllm} & Qwen3-VL-8B/SAM & 45.28 & 39.98 & 34.56 & 30.77 & 11.88 & 39.79 & 47.39\\
ReSaP (8B)~\cite{lv2026resap} & Qwen3-VL-8B/SAM & \underline{51.28} & \underline{47.88} & \underline{43.06} & \underline{36.62} & 19.33 & 45.62 & \underline{53.44}\\
\midrule
\multicolumn{9}{@{}l}{\emph{Generative RGB-rendering models}}\\
\editor & Qwen2.5-VL/DiT & 46.83 & 41.26 & 37.60 & 32.50 & \underline{23.13} & \underline{47.36} & 47.09\\
\method & Qwen2.5-VL/DiT & \textbf{54.83} & \textbf{49.84} & \textbf{46.44} & \textbf{40.75} & \textbf{29.13} & \textbf{52.78} & \textbf{54.01}\\
\bottomrule
\end{tabular}
\end{table*}

\begin{table*}[!t]
\centering
\caption{\textbf{Region-union UAV referring results} (\%). Each \editor--\method pair shares rendered candidates. Best/second-best values per benchmark are bold/underlined.}
\label{tab:region-main}
\footnotesize
\renewcommand{\arraystretch}{1.08}
\setlength{\tabcolsep}{1.8pt}
\begin{tabular}{@{}lllrrrrrrr@{}}
\toprule
Dataset & Method & Core backbone(s) & P@0.5 & P@0.6 & P@0.7 & P@0.8 & P@0.9 & mIoU & oIoU\\
\midrule
\multirow{6}{*}{UAVid-RIS}
 & LAVT~\cite{yang2022lavt} & Swin-B/BERT & 84.45 & 75.75 & 59.60 & 38.71 & 10.91 & 69.32 & 78.76\\
 & RMSIN~\cite{liu2024rmsin} & Swin-B/BERT & 85.71 & 77.91 & 64.06 & 46.52 & \underline{17.76} & 72.05 & \underline{81.10}\\
 & MAFN~\cite{shi2025mafn} & Swin-B/BERT & 85.10 & 77.76 & 63.68 & 44.26 & 16.33 & 71.64 & 80.67\\
 & AeroReformer~\cite{li2025aeroreformer} & Swin-B/BERT & \underline{86.34} & \underline{79.07} & \underline{65.60} & \textbf{47.12} & \textbf{18.52} & \underline{72.79} & \textbf{81.53}\\
 & \editor & Qwen2.5-VL/DiT & 81.37 & 72.26 & 59.24 & 42.23 & 15.05 & 69.82 & 73.01\\
 & \method & Qwen2.5-VL/DiT & \textbf{88.37} & \textbf{79.68} & \textbf{66.24} & \underline{46.82} & 17.24 & \textbf{73.48} & 78.66\\
\midrule
\multirow{6}{*}{VDD-RIS}
 & LAVT~\cite{yang2022lavt} & Swin-B/BERT & 90.04 & 86.72 & 76.01 & 56.09 & 34.32 & 77.80 & 82.51\\
 & RMSIN~\cite{liu2024rmsin} & Swin-B/BERT & 91.14 & 85.98 & 76.38 & 57.20 & 35.06 & 78.22 & 83.58\\
 & MAFN~\cite{shi2025mafn} & Swin-B/BERT & \underline{91.51} & \underline{87.82} & \underline{79.70} & 60.89 & 35.06 & 79.21 & \underline{83.97}\\
 & AeroReformer~\cite{li2025aeroreformer} & Swin-B/BERT & \textbf{92.99} & \textbf{89.30} & \textbf{81.55} & \underline{63.47} & 38.75 & \textbf{80.72} & \textbf{85.38}\\
 & \editor & Qwen2.5-VL/DiT & 86.72 & 81.18 & 74.91 & 62.73 & \textbf{44.65} & 78.63 & 76.64\\
 & \method & Qwen2.5-VL/DiT & 89.67 & 83.39 & 76.38 & \textbf{64.21} & \underline{41.33} & \underline{79.69} & 81.58\\
\bottomrule
\end{tabular}
\end{table*}

\begin{figure*}[!t]
  \centering
  \includegraphics[width=0.985\textwidth]{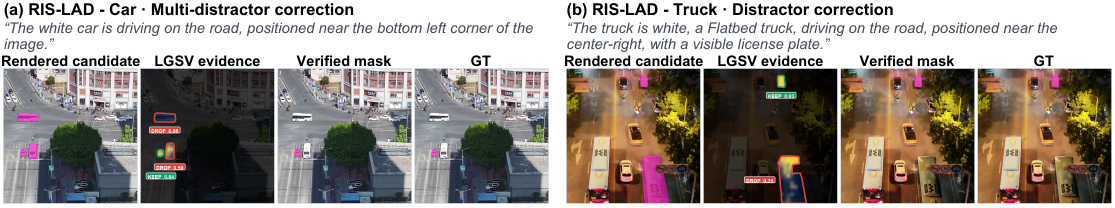}
  \caption{\textbf{Proposal-constrained verification on RIS-LAD.} \captionpanel{a} Car; \captionpanel{b} truck. All panels for each example use the same target-centered visualization crop. The evidence panels display $P\odot M_{\rm raw}$, with contours inherited from the decoded candidates. Green and red contours mark retained and rejected components, respectively; numerical labels show mean component evidence $s_i$.}
  \label{fig:rislad-verification}
\end{figure*}

\begin{figure*}[!t]
  \centering
  \includegraphics[width=0.985\textwidth]{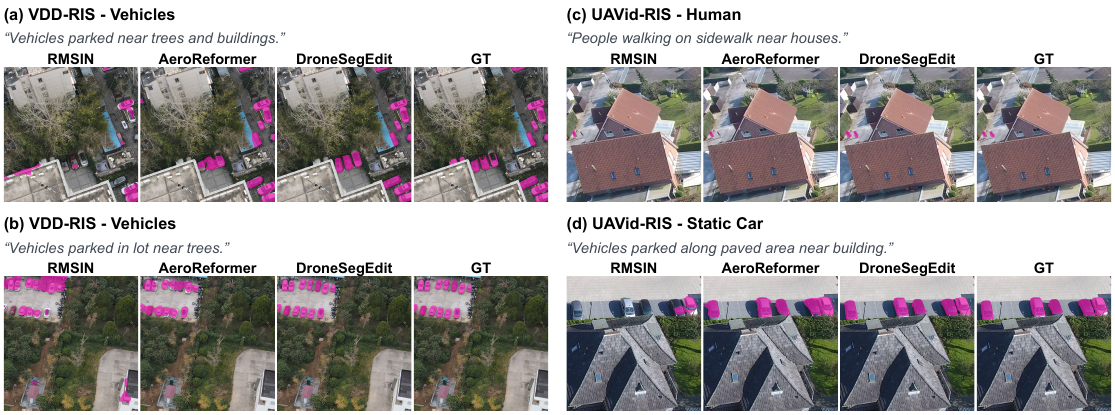}\\[4pt]
  \includegraphics[width=0.86\textwidth]{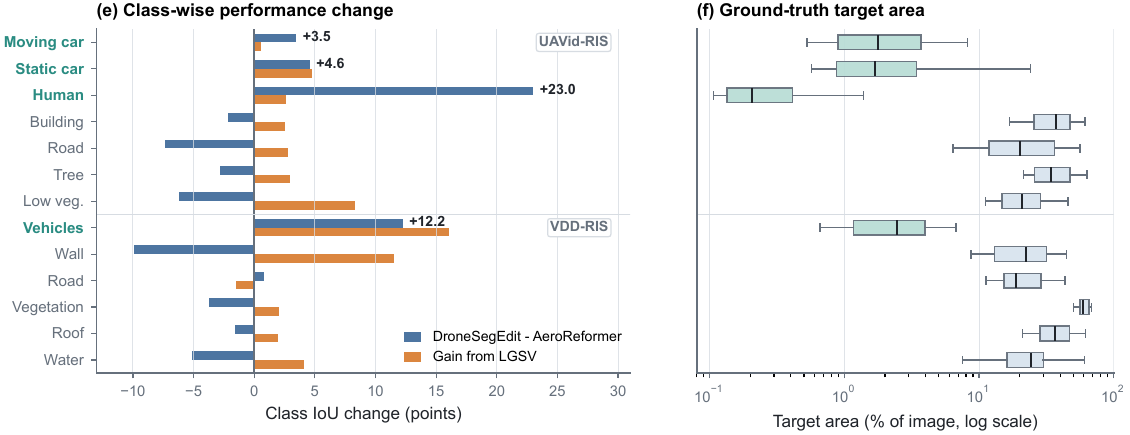}
  \caption{\textbf{Cross-benchmark compact-target evidence.} \captionpanel{a--d} For each UAVid-RIS and VDD-RIS example, all methods are shown using the same target-centered visualization crop, enlarged for readability. \method keeps adjacent vehicle regions separated in \captionpanel{a}, \captionpanel{b}, and \captionpanel{d}, and retains both compact human regions in \captionpanel{c}, more closely matching GT. \captionpanel{e} Class-wise IoU changes. \captionpanel{f} Per-expression target area $100|Y|/(HW)$ on a log scale; boxes show quartiles and 5th--95th-percentile whiskers.}
  \label{fig:cross-benchmark}
\end{figure*}

\subsection{Task Suite and Evaluation}

\paragraph{RIS-LAD.}
RIS-LAD is a particularly demanding single-instance benchmark~\cite{ye2026rislad}.  Across the complete official test set, the referred mask occupies a median of only 0.246\% of the image; 69.9\% and 82.7\% of the targets occupy less than 0.5\% and 1\% of the image area, respectively.  These compact referents frequently appear amid repeated or visually similar vehicles and people, and may be distinguished only through attributes, spatial relations, or scene context.  Its varied illumination, including daytime and nighttime scenes, further changes target appearance.  The benchmark therefore couples compact-object localization with fine-grained instance-level language grounding under substantial appearance variation.  We evaluate on the complete official test split using the benchmark's published metrics; checkpoint and decision settings are selected on validation only.

\paragraph{UAVid-RIS and VDD-RIS.}
UAVid-RIS includes disconnected class-level unions~\cite{li2025aeroreformer,lyu2020uavid}; VDD-RIS provides the analogous setting over VDD~\cite{li2025aeroreformer,cai2025vdd}.  Both follow the official protocols, with fixed settings reported in the appendix.

\paragraph{VDD dense semantic segmentation.}
For dense VDD parsing, a task-specific adapter renders a complete semantic map from one fixed task-level instruction on the official 280/80/40 split~\cite{cai2025vdd}.  The evaluation uses nearest-palette decoding and reports mIoU over the complete label set.

For referring benchmarks, we report P@0.5--P@0.9, mIoU, and oIoU under the official protocols.  P@$t$ denotes the percentage of expressions whose predicted-mask IoU is at least $t$.  mIoU averages mask IoU over expressions, whereas oIoU divides the accumulated intersection over the evaluation set by the accumulated union.  The \editor and \method rows share identical rendered candidates; their difference is verification alone.  Additional diagnostics appear in the appendix.

\subsection{Implementation Details}

We train each benchmark's adapter independently under its official training and evaluation protocol.  Qwen2.5-VL, the VAE, and the 6B base DiT remain frozen.  The rendering LoRA uses rank 32 and $\alpha=8$, is injected into DiT attention and FFN projections, and is optimized with BF16 AdamW.  It contains 47.19M trainable parameters, approximately 0.79\% of the 6B DiT; referring tasks additionally train a 15.79M LGSV, for 62.98M trainable parameters in total.  For scale, this combined trainable count is approximately 1.05\% of the 6B DiT parameter count, and an even smaller fraction of the complete frozen Qwen--DiT--VAE system.  Referring inference uses eight-step FlowMatch Euler sampling with CFG 2.5.  Across independently adapted referring benchmarks, complete adaptation takes around 20--30 NVIDIA A800 GPU-hours per benchmark, depending on the benchmark-specific schedule.  Resolution, batch size, learning rate, checkpoint, and decoder constants are provided in the appendix.

At inference, RIS-LAD uses 8-connected components with a minimum eligible area of 16 pixels and a relative evidence ratio of 0.75.  The region-union settings are $(\tau_t,d_t)=(0.60,16)$ for UAVid-RIS and $(0.30,8)$ for VDD-RIS.  These decisions operate on fixed RGB candidates; the rendering adapter is frozen before paired Raw/LGSV evaluation.

\subsection{Referring Segmentation Results}

\paragraph{RIS-LAD.}
\method reaches 52.78 mIoU and 54.01 oIoU, outperforming the paired editor by 5.42/6.92 points and ReSaP-8B by 7.16/0.57 points.  Gains hold from P@0.5 (46.83$\rightarrow$54.83) through P@0.9 (23.13$\rightarrow$29.13).  Verification raises precision from 52.93 to 62.10 while recall changes only from 81.04 to 80.57, and moves Pred./GT from 1.531 to 1.297, consistent with removing excess foreground.  Figure~\ref{fig:rislad-verification} spans daylight and nighttime conditions.

\paragraph{Region-union and compact-target evidence.}
LGSV adds 3.66/5.65 mIoU/oIoU on UAVid-RIS and 1.06/4.94 on VDD-RIS.  It obtains the best UAVid-RIS mIoU and P@0.5--P@0.7, but its oIoU remains 2.87 points below AeroReformer.  On VDD-RIS, it obtains the best P@0.8 but trails AeroReformer by 1.03 mIoU and 3.80 oIoU; P@0.9 also decreases from 44.65 to 41.33 relative to the raw editor.  Figure~\ref{fig:cross-benchmark} provides matched examples, class-wise changes, and target-area distributions: the qualitative panels show fewer merged vehicle regions and fewer missed compact human regions, while VDD-RIS Vehicles reaches 82.98 IoU, 12.24 points above AeroReformer, and UAVid-RIS Human reaches 49.68 versus 26.68.  These categories are also among the most compact in their respective benchmarks, complementing the small-target-dominated RIS-LAD evidence in Sec.~4.1.

The task topology helps interpret these gains: RIS-LAD requires rejecting same-class distractors to select one instance, whereas region unions must preserve multiple valid regions.  Subset-constrained filtering can trade recall for precision and cannot recover omitted foreground.

\subsection{Beyond Referring Masks: Single-Instruction Full-Scene UAV Parsing}

We evaluate the shared RGB output interface with a separately trained, verifier-free dense VDD adapter (Sec.~3.1).

\begin{figure}[!t]
  \centering
  \includegraphics[width=0.84\columnwidth]{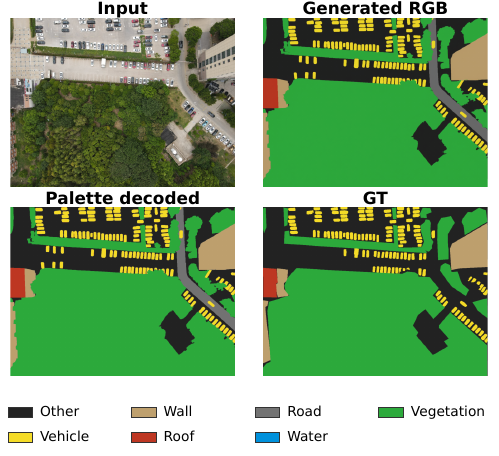}
  \caption{\textbf{Single-instruction dense VDD parsing.} One generation renders the full RGB label map; nearest-palette decoding recovers discrete classes, including densely arranged vehicles, without per-class prompting.}
  \label{fig:vdd-dense}
\end{figure}

At 80.96 mIoU, the verifier-free editor ranks second to UPerNet--Swin-L under the matched-resolution comparison in Table~\ref{tab:vdd-dense}.  Figure~\ref{fig:vdd-dense} shows that a single generation jointly renders the scene label map, while nearest-palette decoding retains the dominant layout and visible object boundaries.  At the selected checkpoint, the mean nearest-palette distance is 11.96, well below the minimum inter-palette separation of 87.61, supporting stable deterministic decoding.

\begin{table}[!t]
\centering
\caption{\textbf{VDD full-scene mIoU} (\%) at matched $1152\!\times\!864$ resolution. All rows use the same split and metric implementation; best/second-best are bold/underlined.}
\label{tab:vdd-dense}
\small
\renewcommand{\arraystretch}{1.04}
\begin{tabular*}{\columnwidth}{@{\extracolsep{\fill}}llr@{}}
\toprule
Method & Backbone & mIoU\\
\midrule
Mask2Former~\cite{cheng2022mask2former} & Swin-T & 74.32\\
Mask2Former~\cite{cheng2022mask2former} & ResNet-50 & 79.90\\
SegFormer-B0~\cite{xie2021segformer} & MiT-B0 & 74.94\\
SegFormer-B5~\cite{xie2021segformer} & MiT-B5 & 78.54\\
UPerNet~\cite{xiao2018upernet} & Swin-T & 80.39\\
UPerNet~\cite{xiao2018upernet} & Swin-L & \textbf{82.17}\\
\method & Qwen2.5-VL/DiT & \underline{80.96}\\
\bottomrule
\end{tabular*}
\end{table}

\section{Analysis}

\begin{table}[!t]
\centering
\caption{\textbf{RIS-LAD validation ablations.} Peak-cell tests whether the highest-evidence cell overlaps target occupancy; Peak-in-mask tests whether its mapped center lies inside the target. Pred./GT is the predicted-to-target area ratio. Boundary IoU and mIoU are in \%.}
\label{tab:ablation}
\small
\setlength{\tabcolsep}{2pt}
\renewcommand{\arraystretch}{1.10}
\begin{tabular*}{\columnwidth}{@{\extracolsep{\fill}}lrrr@{}}
\multicolumn{4}{@{}l}{\textbf{(a) LGSV Design and Language Conditioning}}\\[-1pt]
\toprule
Configuration & Peak-cell & Peak-in-mask & mIoU\\
\midrule
Single layer (L16) & 59.57 & 53.01 & 51.71\\
Mismatched expression & 41.76 & 37.43 & 47.78\\
Equal-capacity image-only & 36.43 & 31.48 & 44.93\\
\method & \textbf{61.48} & \textbf{55.07} & \textbf{53.15}\\
\bottomrule
\end{tabular*}

\vspace{4pt}
\begin{tabular*}{\columnwidth}{@{\extracolsep{\fill}}lrrr@{}}
\multicolumn{4}{@{}l}{\textbf{(b) Generate-and-Verify Component Decomposition}}\\[-1pt]
\toprule
Configuration & Pred./GT & Boundary IoU & mIoU\\
\midrule
LGSV evidence only & 2.489 & 27.68 & 26.12\\
Raw rendered candidate & 1.507 & 53.83 & 50.03\\
Largest component & 1.243 & 50.46 & 48.71\\
\method & \textbf{1.230} & \textbf{58.24} & \textbf{53.15}\\
\bottomrule
\end{tabular*}
\end{table}

Table~\ref{tab:ablation}(a) examines LGSV under fixed rendered proposals.  Four-depth fusion improves over the single-layer readout by 1.44 mIoU and raises Peak-cell/Peak-in-mask by 1.91/2.06 points.  Replacing the matched expression with a mismatched one reduces mIoU from 53.15 to 47.78, while the equal-capacity image-only control reaches only 44.93.  The gain therefore depends on hierarchical, expression-specific evidence rather than added capacity or image saliency alone.

Table~\ref{tab:ablation}(b) separates evidence from geometry.  Thresholding the complete $P$, unlike the proposal-restricted visualization in Fig.~\ref{fig:rislad-verification}, yields only 26.12 mIoU and 27.68 Boundary IoU, with Pred./GT 2.489.  The over-coverage is consistent with coarse occupancy supervision and upsampling.  Thus the supervised readout supplies coarse selection evidence, while rendered proposals provide detailed geometry.  Combining them raises candidate mIoU from 50.03 to 53.15 and Boundary IoU from 53.83 to 58.24.

The largest-component control and LGSV produce similar Pred./GT ratios (1.243 and 1.230), while LGSV improves mIoU by 4.44 points and Boundary IoU by 7.78 points.  This comparison highlights the difference between foreground amount and spatial selection.  Together with the mismatched-expression and image-only controls, it supports the role of expression-conditioned selection within the shared rendered proposals.  Verification cannot recover omitted targets or reliably distinguish candidates with indistinguishable $37\times37$ evidence.

\section{Discussion}

\noindent\textbf{Application-facing evidence.}
In UAV monitoring, unsupported foreground can become a false alarm.  LGSV keeps corrections traceable to editor proposals, and the results support improved selection of compact vehicles and people.

Across all 2,763 RIS-LAD test expressions, verification lowers IoU in 40 cases (1.4\%), with decreases of at least 0.05 IoU in only 20 cases (0.7\%).

\paragraph{Scope and limitations.}
Experiments use \editor with LoRA, and matched controls assess the complementarity of rendering and verification.  New dense label systems require adaptation, and larger palettes may increase color confusion.  Connected color collisions, missed targets, coarse evidence, and reduced-resolution dense evaluation remain limitations.  The large frozen editor and iterative sampling favor offline analysis; reducing memory use and sampling cost is a direction for onboard deployment.

\paragraph{Video and satellite extensions.}
Future work could align evidence across video frames and propagate masks while addressing motion, occlusion, and error accumulation.  Extending the framework to satellite imagery also calls for adaptation to scale and viewpoint shifts.

\section{Conclusion}

\method pairs adapted RGB rendering with proposal-constrained language verification.  It establishes a new state of the art on RIS-LAD and supports compact-target selection on two region-union benchmarks, with mixed aggregate results.  A separately trained dense adapter uses the same RGB interface for full-scene parsing.  Evaluating additional backbones and reducing generation time remain priorities.

\FloatBarrier
\clearpage
{\small
\bibliographystyle{ieeenat_fullname}
\bibliography{references}
}

\clearpage
\newgeometry{margin=0.75in}
\onecolumn
\setcounter{section}{0}
\setcounter{subsection}{0}
\setcounter{figure}{0}
\setcounter{table}{0}
\setcounter{equation}{0}
\renewcommand{\thesection}{\Alph{section}}
\renewcommand{\thesubsection}{\thesection.\arabic{subsection}}
\numberwithin{figure}{section}
\numberwithin{table}{section}
\numberwithin{equation}{section}
\renewcommand{\theHsection}{supp.\arabic{section}}
\renewcommand{\theHsubsection}{supp.\arabic{section}.\arabic{subsection}}
\renewcommand{\theHfigure}{supp.\arabic{section}.\arabic{figure}}
\renewcommand{\theHtable}{supp.\arabic{section}.\arabic{table}}
\renewcommand{\theHequation}{supp.\arabic{section}.\arabic{equation}}
\makeatletter
\renewcommand\section{\@startsection{section}{1}{\z@}%
  {-3.5ex\@plus-1ex\@minus-.2ex}{2.3ex\@plus.2ex}{\normalfont\Large\bfseries}}
\renewcommand\subsection{\@startsection{subsection}{2}{\z@}%
  {-3.25ex\@plus-1ex\@minus-.2ex}{1.5ex\@plus.2ex}{\normalfont\large\bfseries}}
\renewcommand\paragraph{\@startsection{paragraph}{4}{\z@}%
  {3.25ex\@plus1ex\@minus.2ex}{-1em}{\normalfont\normalsize\bfseries}}
\makeatother
\captionsetup{font=small,labelfont=bf,labelsep=period,
  justification=raggedright,singlelinecheck=false,skip=4pt}
\setlength{\parindent}{1em}
\setlength{\parskip}{0pt}
\setlength{\floatsep}{8pt plus 2pt minus 2pt}
\setlength{\textfloatsep}{8pt plus 2pt minus 2pt}
\emergencystretch=2em
\pagestyle{plain}
\raggedbottom
\begin{center}
{\Large\bfseries Appendix\par}
\end{center}
\vspace{0.25em}

\begingroup\small
\noindent\textbf{Scope.}
This appendix provides qualitative evidence, operating limits, and implementation details.  Paired referring comparisons reuse identical rendered RGB candidates, isolating verification; all settings and decisions are fixed before complete-split evaluation.
\par\endgroup

\section{Qualitative Results and Operating Limits}
\label{sec:qualitative-supp}

\subsection{Successful verification}

\begin{figure}[H]
\centering
\includegraphics[width=\suppfigurewidth]{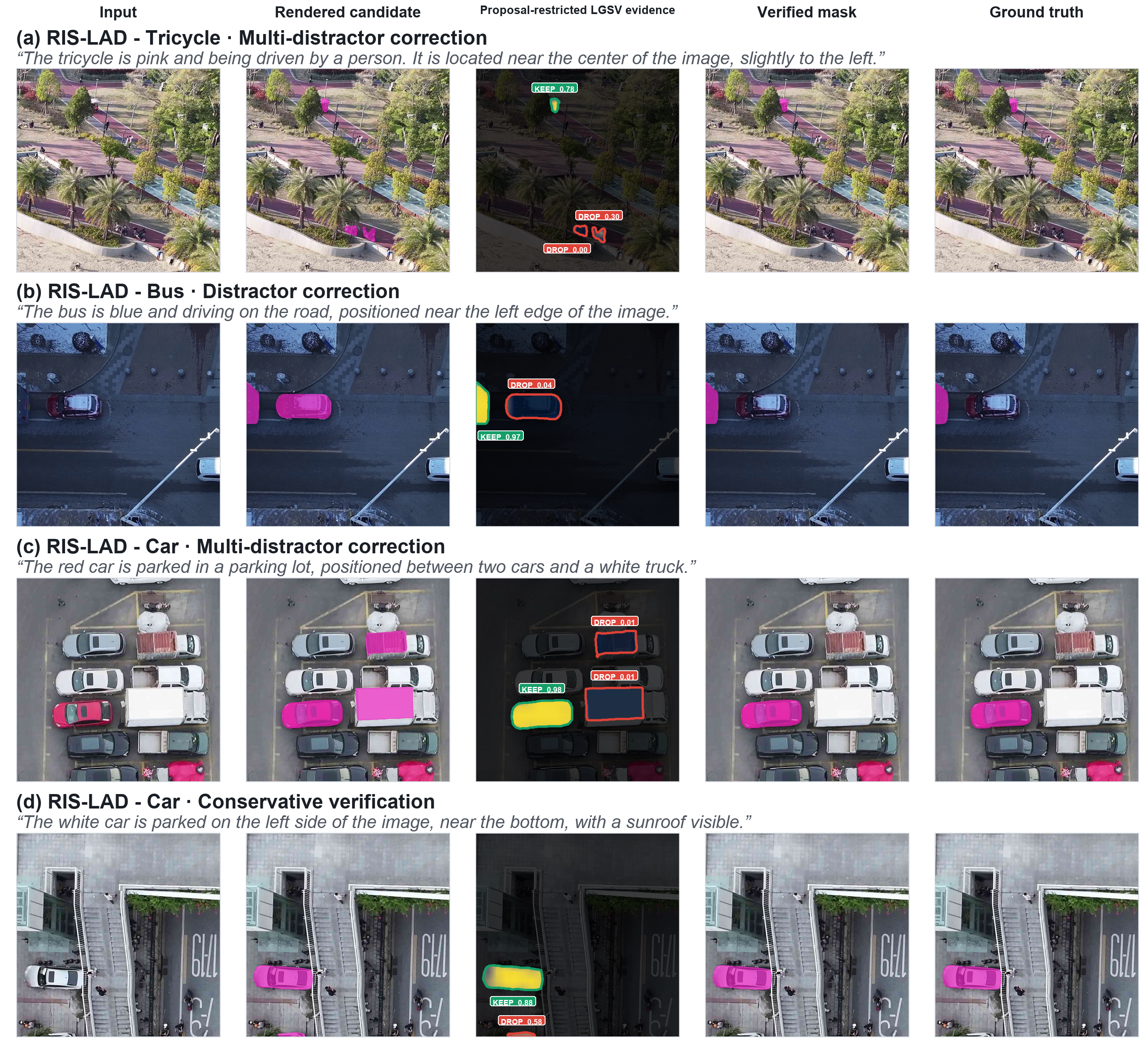}
\caption{\textbf{RIS-LAD verification successes.}  Each row shows the input, fixed RGB-rendered candidate, proposal-restricted LGSV evidence, verified mask, and ground truth.  The evidence panel displays $P\odot M_{\rm raw}$: responses outside decoded candidates are suppressed only for visualization, while its sharp contours come from $M_{\rm raw}$ rather than $P$.  Green/red contours denote retained/rejected components; numerical labels report mean component evidence $s_i$. Panels (a--c) remove unsupported rendering, while (d) conservatively retains an accurate candidate.}
\label{fig:rislad-success-supp}
\end{figure}

Figure~\ref{fig:rislad-success-supp} shows the intended division of labor: the editor renders candidate geometry, while LGSV compares those regions using evidence conditioned on the complete expression.

For RIS-LAD, component scores are computed only over pixels inside each rendered component, and the final mask is constrained to $M_{\rm raw}$.  We therefore show proposal-restricted evidence for decision interpretability; this visualization does not imply that the full evidence map is zero outside the proposals.

\subsection{Region-union benchmark comparisons}

\begin{figure}[H]
\centering
\includegraphics[width=\suppcomparisonwidth]{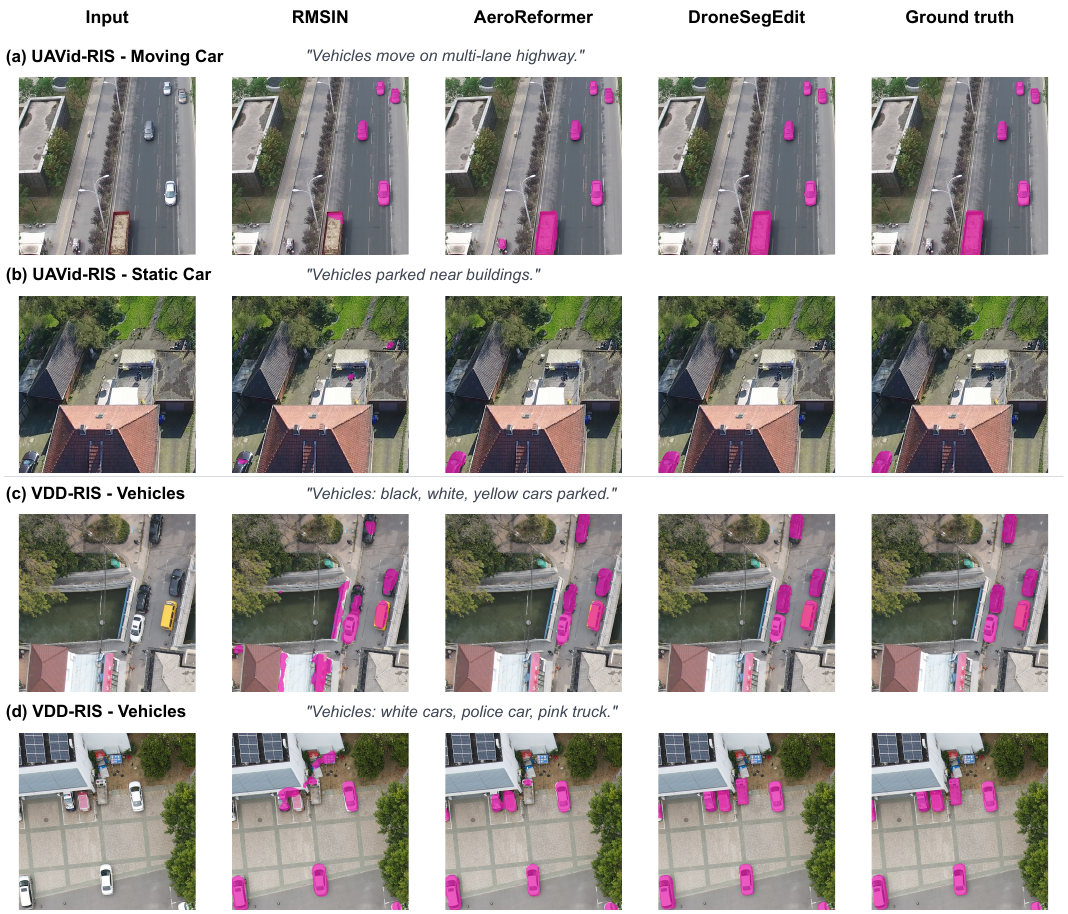}
\caption{\textbf{Region-union benchmark comparisons.}  UAVid-RIS examples (a--b) and VDD-RIS examples (c--d) are shown in one aligned layout.  DroneSegEdit more closely follows the requested moving, parked, and distributed vehicle unions than the comparison methods, while avoiding unrelated foreground.  Every panel is rendered directly from the original image and its corresponding prediction mask; no per-example metric is displayed because complete-split results are reported in the main paper.}
\label{fig:region-union-cross-supp}
\end{figure}

We present UAVid-RIS and VDD-RIS jointly as region-union benchmarks because one expression can refer to multiple disconnected regions.  Figure~\ref{fig:region-union-cross-supp} compares RMSIN, AeroReformer, and DroneSegEdit against the same ground truth using source-resolution crops.  The examples complement the complete-split tables without repeating the main-paper cases.

\subsection{Failure cases and diagnostic frequency}

LGSV is proposal-constrained: $M_{\rm final}\subseteq M_{\rm raw}$.  It can remove unsupported foreground already present in the candidate but cannot recover a target omitted by the editor.  Its coarse evidence can also misrank nearby proposals, and component selection cannot divide an indivisible rendered component.

On all 2,763 RIS-LAD test expressions, verification lowers IoU in 40 cases (1.4\%); only 20 cases (0.7\%) decrease by at least 0.05 IoU.

\begin{figure}[H]
\centering
\includegraphics[width=\suppfigurewidth]{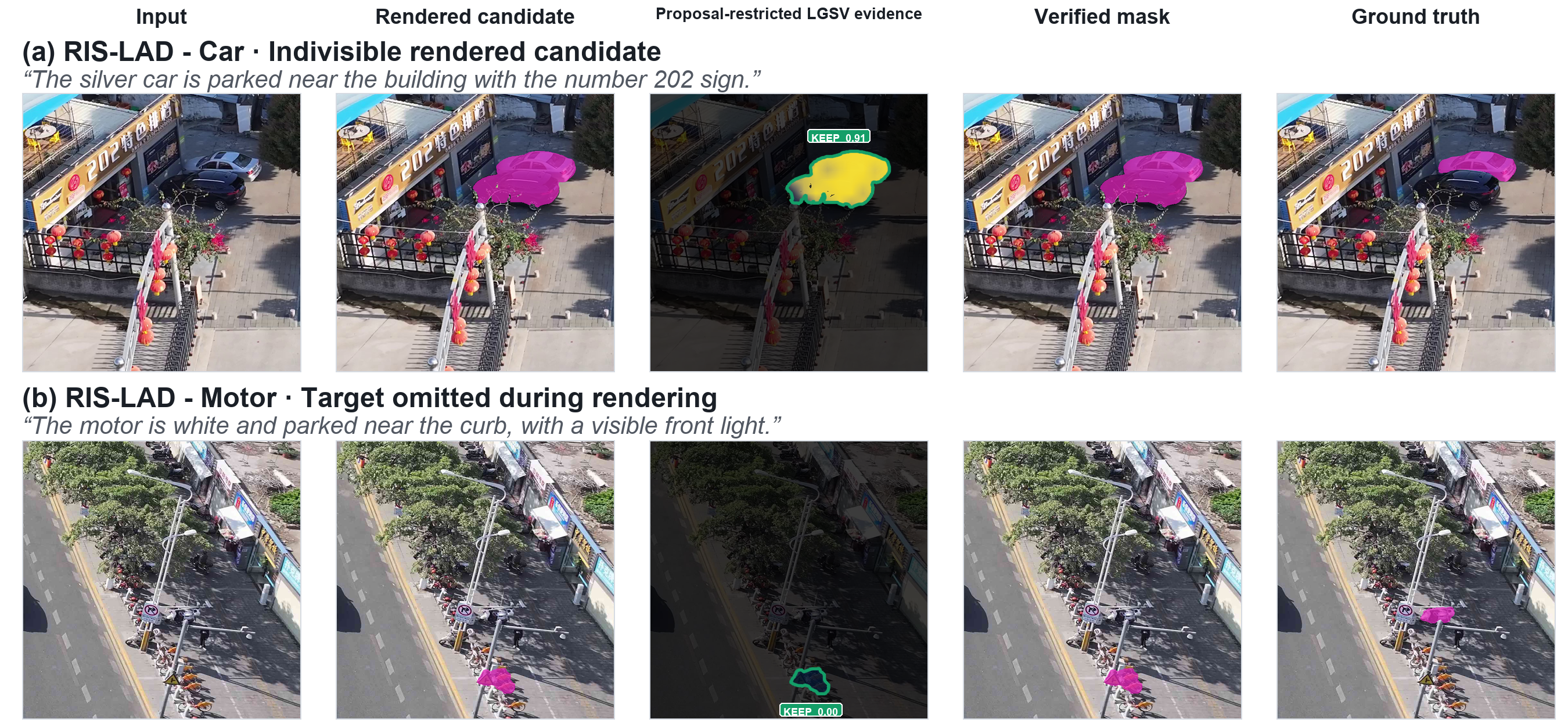}
\caption{\textbf{Representative RIS-LAD operating limits.}  Top: a connected rendering merges the referred car with adjacent vehicles and cannot be divided by component selection.  Bottom: proposal omission leaves no target foreground for LGSV to recover.  Both cases have unambiguous referents and expose the one-sided scope of verification without changing the fixed candidate-generation protocol.}
\label{fig:rislad-failures-supp}
\end{figure}

\section{Implementation Details}

\subsection{Notation}

Table~\ref{tab:notation} fixes notation across the main paper and appendix.  Roman subscripts name semantic roles (e.g., ``raw'' and ``final''), $t$ indexes tasks, $\ell$ indexes frozen MLLM depth, and $x$ indexes pixels.  Calligraphic letters denote sets or operators; bold lowercase symbols denote RGB colors or feature vectors.

\begin{table}[H]
\centering
\caption{Symbols shared by the main paper and appendix.}
\label{tab:notation}
\small
\renewcommand{\arraystretch}{1.04}
\setlength{\tabcolsep}{4.5pt}
\begin{tabular}{>{\raggedright\arraybackslash}p{3.05cm}>{\raggedright\arraybackslash}p{10.35cm}}
\toprule
Symbol & Definition\\
\midrule
$t\in\mathcal T$ & Task index, with $\mathcal T=\mathcal T_{\rm ref}\cup\mathcal T_{\rm sem}$.\\
$\mathcal T_{\rm ref}$ & Referring tasks, partitioned as $\mathcal T_{\rm single}\cup\mathcal T_{\rm union}$.\\
$I,e,Y_t$ & Input RGB image, optional referring expression, and task target.\\
$\Omega,\mathcal C_t$ & Pixel lattice and task label set.\\
$\theta_0,\Delta_t$ & Frozen editor parameters and task-specific LoRA update.\\
$I_g,\mathcal D_{\rm pal}^{(t)}$ & Generated RGB canvas and deterministic task palette decoder.\\
$M_{\rm raw},P,M_{\rm final}$ & Decoded referring candidate, LGSV support map, and verified mask.\\
$F_\ell^v,F_\ell^e,q_\ell,Z_\ell$ & Frozen visual states, expression-token states, pooled expression state, and LGSV logits at depth $\ell$.\\
$(\gamma_\ell,\beta_\ell),\kappa$ & FiLM channel scale/shift and fixed residual modulation scale.\\
$O_\ell$ & Coarse target occupancy at depth $\ell$.\\
$C_i,\mathcal E,\mathcal K_\rho,s_i$ & Rendered component, eligible and retained index sets, and mean evidence score.\\
$\mathcal H_t,\mathcal N_{d_t}(x)$ & Thresholded evidence set and radius-$d_t$ Manhattan neighborhood.\\
\bottomrule
\end{tabular}
\end{table}

\subsection{Data, RGB Interface, and Configurations}
\label{sec:verifier-training-supp}

All reported referring results follow the official evaluation protocols and are computed on the complete official test splits with the benchmark-provided annotations and metrics.  Checkpoints, decoding thresholds, and LGSV decision settings are selected on validation only; test annotations are not used for selection.  Decoded masks are resized to each benchmark's official evaluation resolution using nearest-neighbor interpolation before metric computation.

All tasks use the same frozen \editor base~\cite{longcat2025} with task-specific LoRA adaptation~\cite{hu2022lora}.  For fair comparison, each dataset/task follows its official training and evaluation protocol with an independently trained adapter, avoiding confounding gains from cross-task data mixing.  For an adapted projection $W_j\in\mathbb R^{d_j^{\rm out}\times d_j^{\rm in}}$, task $t$ learns
\begin{equation}
\begin{gathered}
W_j^{(t)}=W_j+\Delta W_j^{(t)},\qquad
\Delta W_j^{(t)}=\frac{\alpha_j}{r_j}B_j^{(t)}A_j^{(t)},\\[-1pt]
\operatorname{rank}(\Delta W_j^{(t)})\le r_j,
\end{gathered}
\label{eq:lora-supp}
\end{equation}
where $B_j^{(t)}\in\mathbb R^{d_j^{\rm out}\times r_j}$ and $A_j^{(t)}\in\mathbb R^{r_j\times d_j^{\rm in}}$.  For referring tasks, the deterministic palette decoder identifies magenta-like pixels from the generated 8-bit channels: red and blue must be high, green must be low, and both red--green and blue--green must exceed a fixed margin.  A pixel is decoded as foreground if and only if
\begin{equation}
R\geq t_R\ \wedge\ B\geq t_B\ \wedge\ G\leq t_G
\ \wedge\ R-G\geq t_\Delta\ \wedge\ B-G\geq t_\Delta.
\label{eq:palette}
\end{equation}
This joint rule rejects bright or gray scene pixels that satisfy only the absolute-channel tests.  The resulting binary mask is $M_{\rm raw}$, and the same thresholds are used for every sample.

\paragraph{RIS-LAD.}
We use the fixed instruction template
\begin{quote}
\small\ttfamily\raggedright Color the only \{target\_class\} described as \char34\{expression\}\char34{} pure magenta. Keep everything else unchanged.
\end{quote}
The category placeholder does not introduce external label information.  The official RIS-LAD formulation obtains its class-level feature by extracting the category token from the referring expression~\cite{ye2026rislad}.  Accordingly, \{target\_class\} is deterministically extracted from the provided expression; no ground-truth category metadata or test annotation is accessed.

\paragraph{UAVid-RIS and VDD-RIS.}
Similarly, for the two region-union benchmarks, we use
\begin{quote}
\small\ttfamily\raggedright Paint the \{class\_name\} region matching \char34\{expression\}\char34{} with pure magenta.
\end{quote}

\paragraph{Dense VDD.}
We deliberately use a simple task-level instruction:
\begin{quote}
\small\ttfamily\raggedright Convert the full image into an RGB semantic label map.
\end{quote}
The prompt contains neither class names nor palette-color constraints.  Instead, the complete class-to-color correspondence and joint full-scene parsing behavior are learned from paired RGB targets during task-specific LoRA SFT.  Thus language specifies only the task, while supervision supplies the label inventory and output protocol.  All templates are fixed before evaluation, and no per-example prompt selection is performed.

\paragraph{Shared settings.}
All LoRAs use rank 32 and $\alpha=8$ in the DiT self-/cross-attention Q/K/V/output and self-/context-FFN projections; Qwen2.5-VL, the base DiT, and the VAE remain frozen.  Rendering LoRAs use BF16 AdamW, weight decay $10^{-2}$, uniform diffusion-timestep sampling, and no stochastic image augmentation.  Referring inference uses a $1024^2$ canvas, FlowMatch Euler sampling with dynamic shifting, 8 steps, CFG 2.5, and $(t_R,t_B,t_G,t_\Delta)=(190,170,100,30)$.  LGSV is optimized separately over frozen Qwen2.5-VL states using BF16 AdamW, batch 2, and weight decay $10^{-2}$.  It uses layers $\{8,16,24,28\}$ and width 256; its fixed settings are $(\lambda_b,\lambda_d,\lambda_c)=(1,1,0.50)$, $\kappa=0.10$, positive-weight cap $\omega_{\max}=50$, and Dice smoothing $\epsilon=1$.  Region-union tasks omit only the center-cell term.  LoRA contains 47.19M optimized parameters, increasing to 62.98M when the 15.79M LGSV heads are trained for a referring task.

\begin{table}[H]
\centering
\caption{\textbf{Training and inference configurations.} Rendering-LoRA schedules and checkpoints are listed separately from LGSV optimization.  LoRA checkpoints determine the fixed RGB candidates, whereas LGSV checkpoints are selected independently on validation.  RIS-LAD uses consecutive constant-LR adaptation and lower-LR cosine-decay continuation stages; all other LoRA schedules are single-stage.}
\label{tab:hyperparams-supp}
\footnotesize
\renewcommand{\arraystretch}{1.10}
\setlength{\tabcolsep}{3.0pt}
\begin{tabular}{>{\raggedright\arraybackslash}p{2.25cm}>{\raggedright\arraybackslash}p{2.95cm}>{\raggedright\arraybackslash}p{3.25cm}>{\raggedright\arraybackslash}p{2.45cm}>{\raggedright\arraybackslash}p{3.75cm}}
\toprule
Module / task & Optimization & Schedule & Selected checkpoint & Inference\\
\midrule
Rendering LoRA: RIS-LAD & AdamW; LR $10^{-4}$; batch 4. Continuation: $5\!\times\!10^{-5}$ & 2k constant-LR adaptation; 2k cosine-decay continuation & 2k adaptation + 2k continuation (val.) & $1024^2$; 8 steps; CFG 2.5; fixed decoder\\
Rendering LoRA: UAVid-RIS & AdamW; LR $10^{-4}$; batch 8 & 7k constant & 6k (val.) & $1024^2$; 8 steps; CFG 2.5; fixed decoder\\
Rendering LoRA: VDD-RIS & AdamW; LR $10^{-4}$; batch 8 & 7k constant & 5k (val.) & $1024^2$; 8 steps; CFG 2.5; fixed decoder\\
Rendering LoRA: VDD dense & AdamW; LR $10^{-4}$; batch 2 & 7k constant & 5k (val.) & $1152\!\times\!864$; 12 steps; CFG 1.5; nearest palette; no LGSV\\
\midrule
LGSV readout: referring tasks & BF16 AdamW; batch 2; weight decay $10^{-2}$; LR $10^{-4}$ & At most 24k optimizer updates & Best on validation & Layers $\{8,16,24,28\}$; width 256\\
\bottomrule
\end{tabular}
\end{table}

Complete adaptation takes around 20 NVIDIA A800 GPU-hours on RIS-LAD and around 30 GPU-hours on each of UAVid-RIS and VDD-RIS.  GPU-hours denote the number of GPUs multiplied by training wall-clock time for one benchmark-specific run.

Across RIS-LAD, UAVid-RIS, and VDD-RIS, LGSV shares the same readout layers, width, FiLM parameterization, optimizer family, batch size, weight decay, and BCE/Dice weights.  RIS-LAD additionally activates the center-cell term, while each benchmark retains its validation-selected LGSV checkpoint and output-topology decision rule.  The fixed decision settings are: RIS-LAD uses $A_{\min}=16$, mean component evidence, $\rho=0.75$, and 8-connectivity; UAVid-RIS uses $(\tau_t,d_t)=(0.60,16)$; and VDD-RIS uses $(\tau_t,d_t)=(0.30,8)$.

During integrated inference, LGSV reuses the multi-level hidden states already produced by the editor's Qwen conditioning pass, so verification requires no second Qwen forward.  LGSV training and cached-output diagnostics instead run the frozen Qwen encoder separately for offline feature extraction.

The rendering-LoRA checkpoint is frozen before paired Raw/LGSV evaluation.  LGSV therefore changes only the language-conditioned verification readout and does not resample or update the RGB candidate generator.  Dense VDD uses the rendering LoRA alone.

\subsection{LGSV Supervision Details}

The main paper specifies the verifier architecture, loss, and shared subset operator.  Here we make the coarse supervision target explicit.  At depth $\ell$, adaptive max pooling maps the task target mask $Y_t$ to the evidence grid:
\begin{equation}
O_\ell=\mathbf 1[\operatorname{AdaMaxPool}_\ell(Y_t)>0].
\label{eq:coarse-targets}
\end{equation}
Here $O_\ell$ is target occupancy.  Full-grid BCE labels non-target cells as negatives under expression conditioning, while Dice penalizes excessive predicted support.  The BCE positive weight uses the background-to-target occupancy ratio capped by $\omega_{\max}$, and the center term uses the target-centroid cell only for RIS-LAD.  Adaptive max pooling intentionally supplies occupancy rather than downsampled contours: LGSV learns coarse expression support, while the rendered candidate supplies the proposal geometry.  The fixed coefficients are listed above Table~\ref{tab:hyperparams-supp}.

\subsection{Ablation Protocol Details}

All ablations follow the same RIS-LAD validation protocol.  Matched LGSV design comparisons reuse fixed rendered proposals.  The mismatched-expression control disrupts language--target correspondence, whereas the equal-capacity image-only control removes expression modulation while retaining readout capacity.  The evidence-only control directly uses the validation-fixed evidence map without a rendered proposal, and the largest-component control retains the largest eligible decoded component.

\section{Additional Diagnostics}

\subsection{Single-Instruction Full-Scene VDD Parsing}

A separately trained dense VDD adapter receives a complete UAV image and jointly renders seven labels at $1152\times864$, preserving the native $4{:}3$ aspect ratio~\cite{cai2025vdd}.  The fixed instruction in Sec.~\ref{sec:verifier-training-supp} contains neither class names nor colors; paired RGB targets teach their correspondence, without per-class prompt enumeration or repeated class-specific calls.

\begin{figure}[H]
  \centering
  \includegraphics[width=\suppfigurewidth]{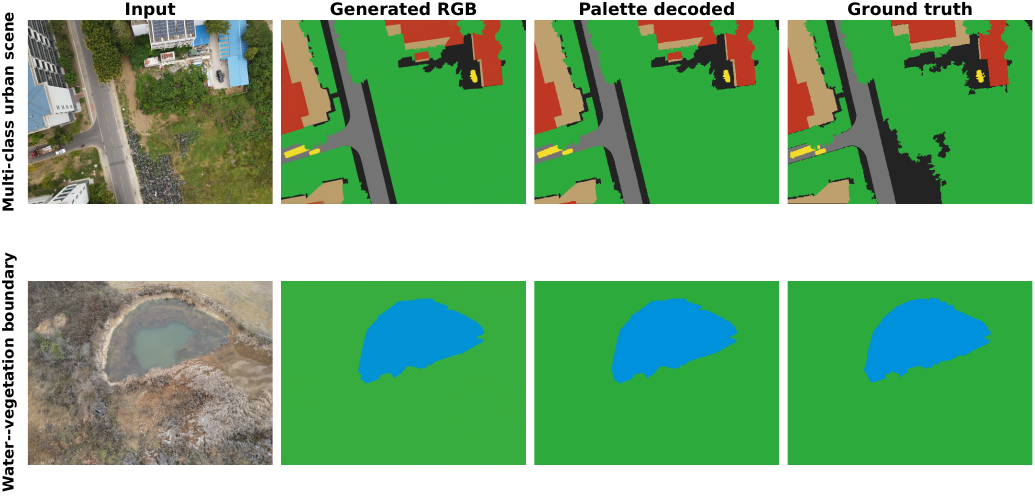}
  \caption{\textbf{Single-instruction dense VDD outputs.} The adapter renders all classes jointly as RGB, and nearest-palette decoding maps the continuous output to discrete labels. The two rows illustrate a multi-class urban scene and a water--vegetation boundary.}
  \label{fig:vdd-palette-qualitative}
\end{figure}

\begin{figure}[H]
  \centering
  \includegraphics[width=0.78\textwidth]{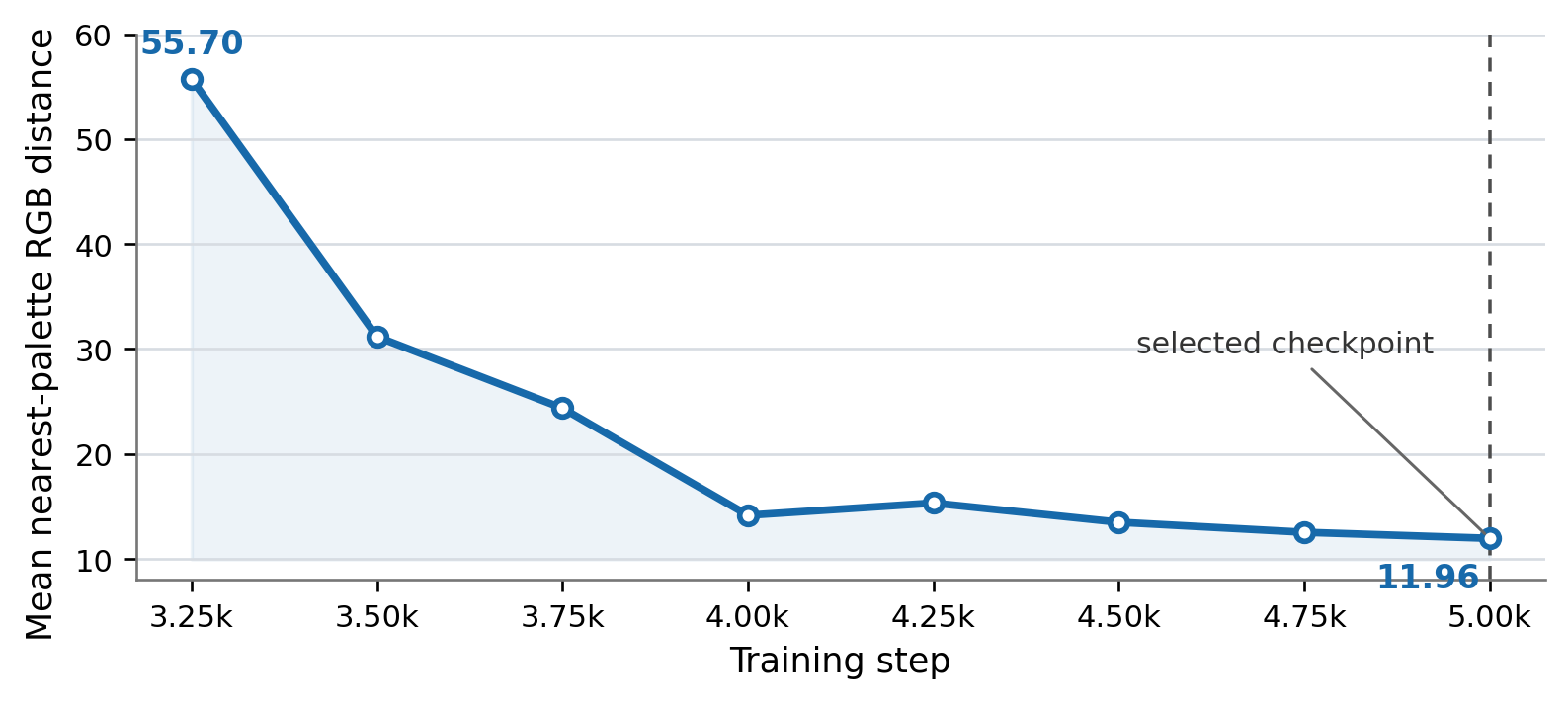}
  \caption{\textbf{Nearest-palette distance during dense VDD adaptation.} The selected checkpoint lies in the converged low-distance regime; the final mean distance (11.96) remains far below the minimum inter-palette separation (87.61).}
  \label{fig:vdd-palette-convergence}
\end{figure}

To measure acquisition of this implicit output protocol, we compute the distance from each generated pixel to its nearest valid palette color,
\begin{equation}
d_{\rm pal}(I_g)=\frac{1}{|\Omega|}\sum_{x\in\Omega}
\min_{\mathbf c\in\mathcal C_{\rm pal}}\lVert I_g(x)-\mathbf c\rVert_2.
\label{eq:palette-distance}
\end{equation}
At the selected checkpoint, the mean distance is 11.96 in 8-bit RGB space, while the minimum distance between any two palette colors is 87.61.  Generated pixels are, on average, substantially closer to a valid palette entry than the minimum inter-palette separation, supporting stable nearest-palette decoding. Figures~\ref{fig:vdd-palette-qualitative} and~\ref{fig:vdd-palette-convergence} show representative outputs and the convergence of this diagnostic.

For reproducibility, the ordered palette is
\begin{center}
\setlength{\tabcolsep}{5pt}
\renewcommand{\arraystretch}{1.12}
\begin{tabular}{@{}cll@{\hspace{2.5em}}cll@{}}
$\mathbf c_0$ & other      & $(35,35,35)$    & $\mathbf c_1$ & wall       & $(190,160,110)$\\
$\mathbf c_2$ & road       & $(115,115,115)$ & $\mathbf c_3$ & vegetation & $(45,170,60)$\\
$\mathbf c_4$ & vehicle    & $(245,220,40)$  & $\mathbf c_5$ & roof       & $(190,55,35)$\\
$\mathbf c_6$ & water      & $(0,145,220)$   &               &            &\\
\end{tabular}
\end{center}
The generated RGB image is decoded by nearest Euclidean distance to these colors, with the lower class ID used only to break an exact tie.  All seven decoded classes, including \emph{other}, enter the reported mIoU; no pixel is ignored or left unassigned.

Under the complete seven-class protocol at $1152\!\times\!864$ input, the selected checkpoint obtains 80.96 mIoU.  This verifier-free experiment evaluates the shared RGB output interface.  The dense adapter is trained separately, without transferring the trained referring adapter or LGSV.  It remains a closed-set, task-specific adaptation: changing the label system requires retraining or adaptation, and larger label sets may introduce palette crowding or color confusion.  A frozen-VAE reconstruction check reaches 99.296 mIoU on 64 validation images after the same palette readout, indicating limited reconstruction-and-decoding loss on this validation subset.

\subsection{Complementary Foreground Diagnostics}

\begin{table}[H]
\centering
\caption{Complementary diagnostics not repeated in the main comparison tables.  Each \editor--\method pair reuses the same rendered RGB candidate.  mDice is averaged over expressions, whereas global Precision, Recall, and Pred/GT pool pixels over the evaluation set.  Values are percentages except Pred/GT, for which values closer to one are preferred.}
\label{tab:thresholds}
\small
\renewcommand{\arraystretch}{1.06}
\setlength{\tabcolsep}{6pt}
\begin{tabular}{@{}llrrrr@{}}
\toprule
Dataset & Output & mDice & \begin{tabular}[c]{@{}c@{}}Global\\Precision\end{tabular} & \begin{tabular}[c]{@{}c@{}}Global\\Recall\end{tabular} & Pred/GT\\
\midrule
\multirow{2}{*}{RIS-LAD}
&\editor&54.37&52.93&\textbf{81.04}&1.531\\
&\method&\textbf{58.88}&\textbf{62.10}&80.57&\textbf{1.297}\\
\midrule
\multirow{2}{*}{VDD-RIS}
&\editor&85.69&80.44&\textbf{94.20}&1.171\\
&\method&\textbf{86.90}&\textbf{87.70}&92.11&\textbf{1.050}\\
\midrule
\multirow{2}{*}{UAVid-RIS}
&\editor&80.04&76.93&\textbf{93.47}&1.215\\
&\method&\textbf{83.05}&\textbf{85.84}&90.38&\textbf{1.053}\\
\bottomrule
\end{tabular}
\end{table}

Across all three paired evaluations, verification raises global precision and moves Pred/GT toward one, with a small-to-moderate global-recall trade-off.  These diagnostics support the main paper's interpretation that LGSV primarily removes unsupported rendered foreground from a fixed candidate.

\subsection{Scope of Deployment and Future Extensions}

\paragraph{Model scale and deployment.}
The referring system combines a 47.19M-parameter rendering LoRA with a 15.79M-parameter LGSV (62.98M trainable parameters total), while Qwen2.5-VL, the 6B base DiT, and the VAE remain frozen.  Referring inference uses eight sampling steps with CFG 2.5.  Integrated LGSV reuses the editor's Qwen conditioning states, avoiding an additional Qwen forward.  The frozen backbone and iterative sampling favor offline analysis.  Model compression and fewer-step sampling are directions for onboard deployment.

\paragraph{Temporal UAV imagery.}
The current method processes individual images.  A video extension could align consecutive evidence maps and propagate verified masks using motion or feature correspondences while retaining the per-frame proposal constraint.  Sequence-level evaluation would assess motion, viewpoint changes, occlusion, identity switches, and error accumulation.

\paragraph{Satellite imagery.}
Extending the RGB rendering interface to satellite imagery is another direction for future work.  Differences in ground sampling distance, viewpoint, object scale, and referring expressions motivate task-specific adaptation and evaluation.

\section{Responsible Use}
\label{sec:responsible-use}

The intended scope is civilian research on public UAV benchmarks.  The method is not intended for military targeting or weapon guidance, coercive surveillance, biometric identification, or autonomous high-stakes decisions.  Any deployment requires legal review, data minimization, human oversight, and task-specific validation; third-party data and models remain governed by their original licenses.

\end{document}